\documentclass[runningheads]{llncs}
\usepackage{subfig}
\usepackage{graphicx,wrapfig,lipsum}
\usepackage{graphicx}
\usepackage{algorithmic}
\usepackage{algorithm}
\begin{document}
\title{Dual Active Sampling on Batch-Incremental Active Learning}
%
%
\author{Johan Phan \and
Massimiliano Ruocco\ \and
Francesco Scibilia}
\authorrunning{P. Johan et al.}
%
\institute{Norwegian University of Science and Technology, Trondheim, Norway}
%
%
\maketitle
\begin{abstract}
Recently, Convolutional Neural Networks (CNNs) have shown unprecedented success in the field of computer vision, especially on challenging image classification tasks by relying on a universal approach, i.e., training a deep model on a massive dataset of supervised examples. 
While unlabeled data are often an abundant resource, collecting a large set of labeled data, on the other hand, are very expensive, which often require considerable human efforts. One way to ease out this is to effectively select and label highly informative instances from a pool of unlabeled data (i.e., active learning).
This paper proposed a new method of batch-mode active learning, Dual Active Sampling(DAS),  which is based on a simple assumption,  if two deep neural networks (DNNs) of the same structure and trained on the same dataset give significantly different output for a given sample, then that particular sample should be picked for additional training. 
 While other state of the art methods in this field usually require intensive computational power or relying on a complicated structure, DAS is simpler to implement and, managed to get improved results on Cifar-10 with preferable computational time compared to the core-set method.  
\keywords{Active Learning  \and Deep Learning \and Image Classification.}
\end{abstract}
\section{Introduction}
Over the last few years, Deep Convolutional Neural Networks(CNNs) have completely dominated the field of Image Recognition and proven itself to be a versatile and robust tool for achieving top performance on many tasks. However, as a data-driven method, it requires a considerable amount of labeled data in order to provide a good result.  More importantly, the performance of CNNs are in most cases better with more data, which led to a constant desire to collect more data, even though data labeling is a time consuming and expensive task.
Aiming at improving the performance of an existing model by incrementally selecting and labeling the most suitable/informative unlabeled samples, Active Learning (AL) has been well studied over recent decades, and most of the early work can be found in \cite{activelearningbook}. With the rise in popularity of Deep Learning, especially in using CNNs to solve challenging image recognition problems \cite{ILSVRC15}, several attempts to develop an effective AL strategy on this field have been made, notably Core-set selection\cite{sener2018active}, where the active learner learn to pick the most representative data by treating the problem as a metric k-center. However,  considering that the nature of DNNs are often complex and unpredictable, nearly all of the state of the art methods are often depended on the extracted output information from the networks as the selecting criteria, e.g., Wang et al. \cite{cost2}. While this approach is proven to be effective, it has made the traditional serial AL, i.e., queries and re-train one at a time, less desirable. One of the main reasons is that DNNs are often slow to train and computationally expensive. For this reason, \textbf{batch-mode AL}, i.e. agent queries multiple samples at once, has become a much more suitable approach. 
 As for batch-mode AL, one of the main challenges compared with the serial mode is the lack of constant feedback from the model for each selection, which often leads to overlapping of information between samples in a batch \cite{activelearningbook}, i.e., the majority of them has similar features. Consequently, the goal of the strategies involving batch-mode AL is not to select the "best" sample but to pick the best combination of samples, i.e., the most informative dataset.

This paper proposed a new method of batch-mode AL,  which is based on a simple assumption,  if two DNNs of the same structure and trained on the same dataset give significantly different output for a given sample, then that particular sample should be picked for additional training. While other state of the art methods in this field normally require intensive computational power\cite{sener2018active} or having a complicated structure \cite{cost2}, the method proposed in this paper is fast and simple to implement.
\section{Related work}
When it comes to mapping the landscape of AL, the work "active learning literature survey" by \cite{activelearningbook} has to be mentioned. This paper has addressed all of the basic and advanced methods in AL up until 2010. However, the landscape of AL has changed a lot in the last few years. Followed by the rise of deep learning, most of the focus in AL has switched to supporting DNNs. Some of the notable recent works that put special focus in batch-mode AL for DNNs is,  \cite{cost2} and \cite{ravi2018meta-learning}. The former paper addressed the use of pseudo labelling, and the latter introduces a way to combine batch-mode AL with meta-learning. 

In this work, core-set selection \cite{sener2018active} was chosen as the baseline beside random sampling. This method uses an upper bound of the core-set loss, which is the gap between the training loss on the whole set and the core-set. By minimizing this upper bound, the authors show that the problem is equivalent to a K-center problem which can be solved by using a greedy approximation method or a mixed integer program (MIP) solver. Core-set selection, is on of the current state of the art methods in the field of image classification for CNNs.  However, both of them are very time consuming and could take several days to find the optimal solution, as for core-set selection is to solve an NP-Hard problem that grows exponentially with the size of the data set. 
\section{Theory and Method}

In this paper, the proposed method is called  \textbf{dual active sampling}, since it uses two DNNs with the same structure to perform the selection.
\textbf{Dual Active Sampling (DAS)} or just dual sampling is motivated by the fact that a neural network trained on the same dataset tends to give slightly different results on different runs. In DAS, both DNNs uses the same structure and the same optimization parameters. However, due to data augmentation of the training data, using random image crop and random horizontal flip \cite{DBLP:journals/corr/abs-1811-09030} combined with random dropout, after some training epoch, the internal weights of these networks can become significantly different from each other.  
\begin{algorithm}[tb]
   \caption{Dual Active Sampling(Simple)}
   \label{alg:example}
\begin{algorithmic}
   \STATE Initialize $model1$ and $model2$ with pre-trained weight from Image-Net.
   \STATE Initialize an empty Train set, S and unlabelled dataset, U
   \FOR{$step = 1$ {\bfseries to} $step = N$}
   \IF{$step < M$}
   \STATE Query and Update S with $n$ randomly selected samples from U
    \ELSE
      \FOR{$i = 1$ {\bfseries to} $n$}
     \STATE random pick a batch of sample, \textbf{R} from U
     \STATE index = argmax(distance(model1(R),model2(R))
     \STATE get label for R[index] and add to S 
      \ENDFOR
    \ENDIF
   \FOR{$epoch = 1$ {\bfseries to} $epoch = m $}
   \STATE train(model1)
   \STATE train(model2)
   \STATE validate model1
   \ENDFOR
   \ENDFOR
\end{algorithmic}
\end{algorithm}
\section{Experimental}
\subsection{Datasets and experiment settings}

 Because of time limitation, the experiment was only performed on Cifar-10\cite {Cifar10}, which consists of $60 000$ 32x32 color images from 10 different classes. This dataset was then further divided into a training-set of $48 000$ images, a validation set of $2000$ images and a test-set of $10 000$ images. In the experiment, only samples from the training-set got queried for labels. Additionally, the VGG-16 net had been chosen as the base model because of its simplicity, relative short training time, and powerful performance. The experiment was conducted on PyTorch, where the performance of DAS got compared with random selection and the core-set selection. 
\subsection{Implementation Details}
The algorithm for the experiments can be found in \ref{alg:example}.
The experiments was performed on CIFAR10 using VGG16 with ADAM optimizer \cite{DBLP:journals/corr/KingmaB14}. The learning rate for both networks was chosen to be 0.0001, and only one of the networks got access to the validation-set and got tested on the test-set in order to save time. Both validation-set and test-set in the experiments were fully labeled. Additionally, the Euclidean distance was used to calculate the normalized distance between the output. 

For every step, the agent added n = 100 of labeled data to the training set and trained its networks for m = 10 epochs. In the first two steps, the selection was performed randomly, while for the rest, the agent picked the sample that has the highest distance between networks from a pool of randomly selected samples. The selection was repeated for N = 100 times 
This experiment was performed incremental, i.e., on each new step, the models got trained on the top of the previous step.
\section{Result}
In this section, the accuracy of DAS got compared with random sampling and core-set, both in terms of total accuracy and accuracy per class. The "full-set acc" in Figure 1 is the accuracy of the model that trained on the whole training dataset.

\begin{figure}[ht]
\vskip -0.2in

	   \centering
	   \includegraphics[width=0.5\textwidth]{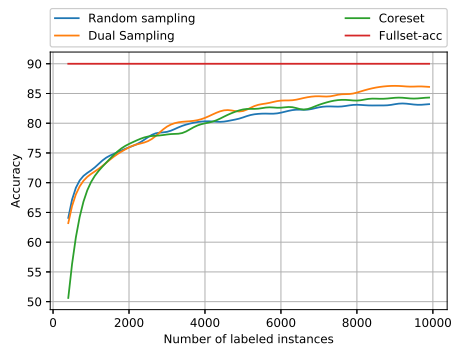}
	   
  \caption{Average test accuacy for Core-set, random sampling and DAS on 9 runs}
  \label{fig:1a}
\vskip -0.4in
\end{figure}

\begin{figure}[!ht]
\vskip -0.4in
  \subfloat[Random Sampling]{
	\begin{minipage}[c][1\width]{
	   0.29\textwidth}
	   \centering
	   \includegraphics[width=1\textwidth]{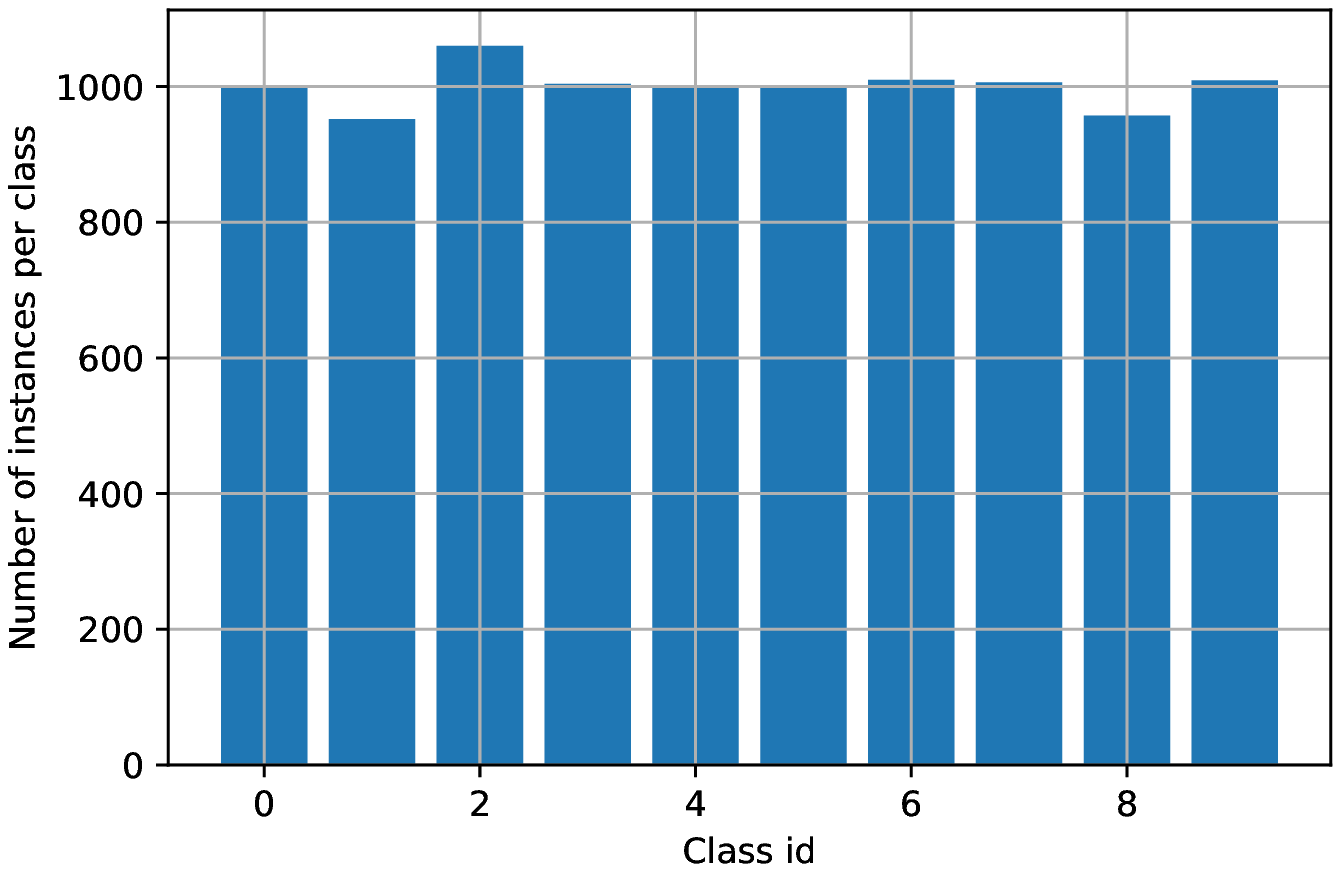}
	\end{minipage}}
 \hfill 	
  \subfloat[Core-set Sampling]{
	\begin{minipage}[c][1\width]{
	   0.29\textwidth}
	   \centering
	   \includegraphics[width=1\textwidth]{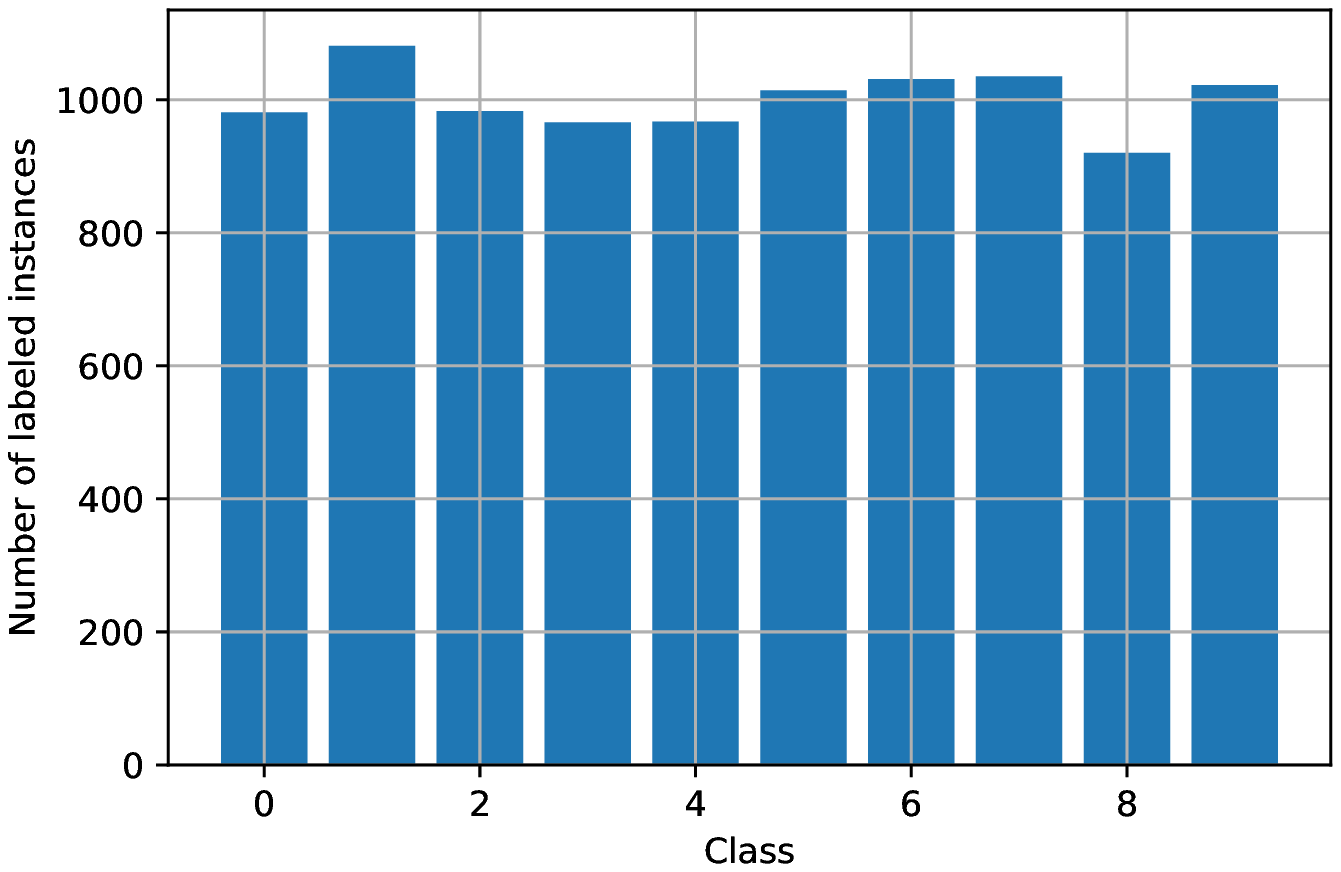}
	\end{minipage}}
 \hfill 	
  \subfloat[DAS]{
	\begin{minipage}[c][1\width]{
	   0.29\textwidth}
	   \centering
	   \includegraphics[width=1\textwidth]{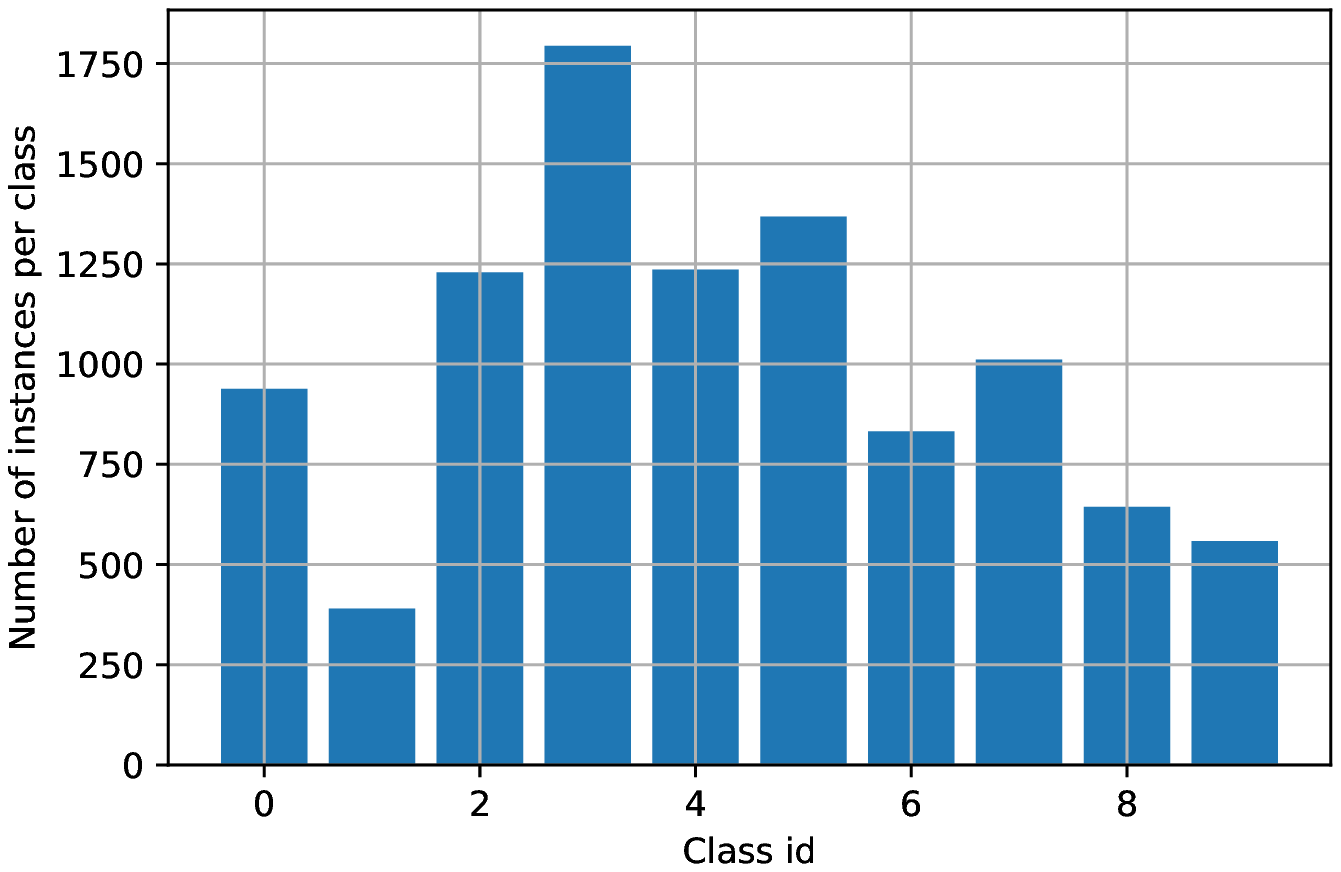}
	\end{minipage}}
\caption{Distribution of items/class for 10000 samples.}

\end{figure}

\begin{figure}[ht]
\vskip -0.4in
  \subfloat[Random Sampling]{
	\begin{minipage}[c][1\width]{
	   0.29\textwidth}
	   \centering
	   \includegraphics[width=1\textwidth]{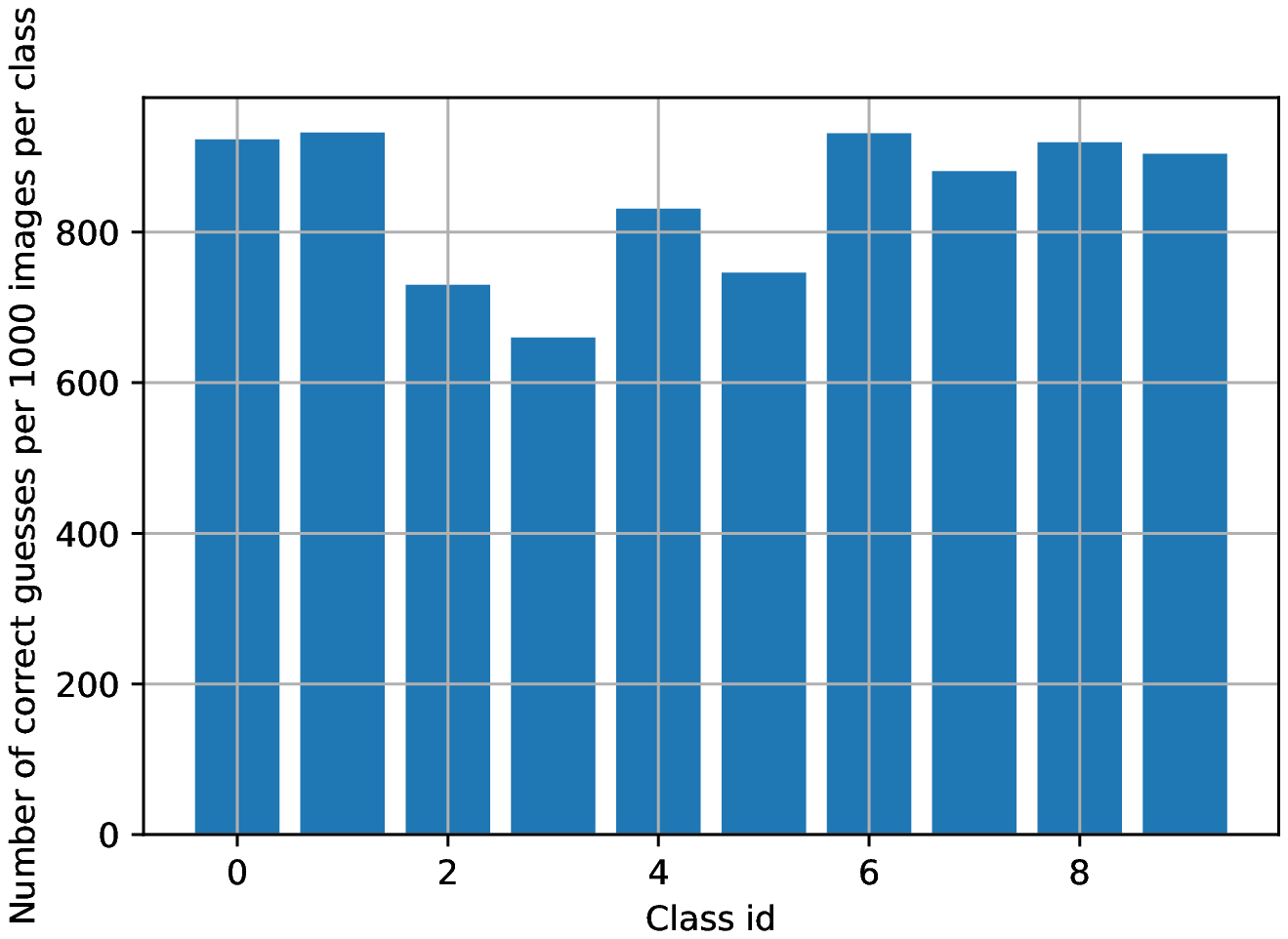}
	\end{minipage}}
 \hfill	
  \subfloat[Core-set Sampling]{
	\begin{minipage}[c][1\width]{
	   0.29\textwidth}
	   \centering
	   \includegraphics[width=1\textwidth]{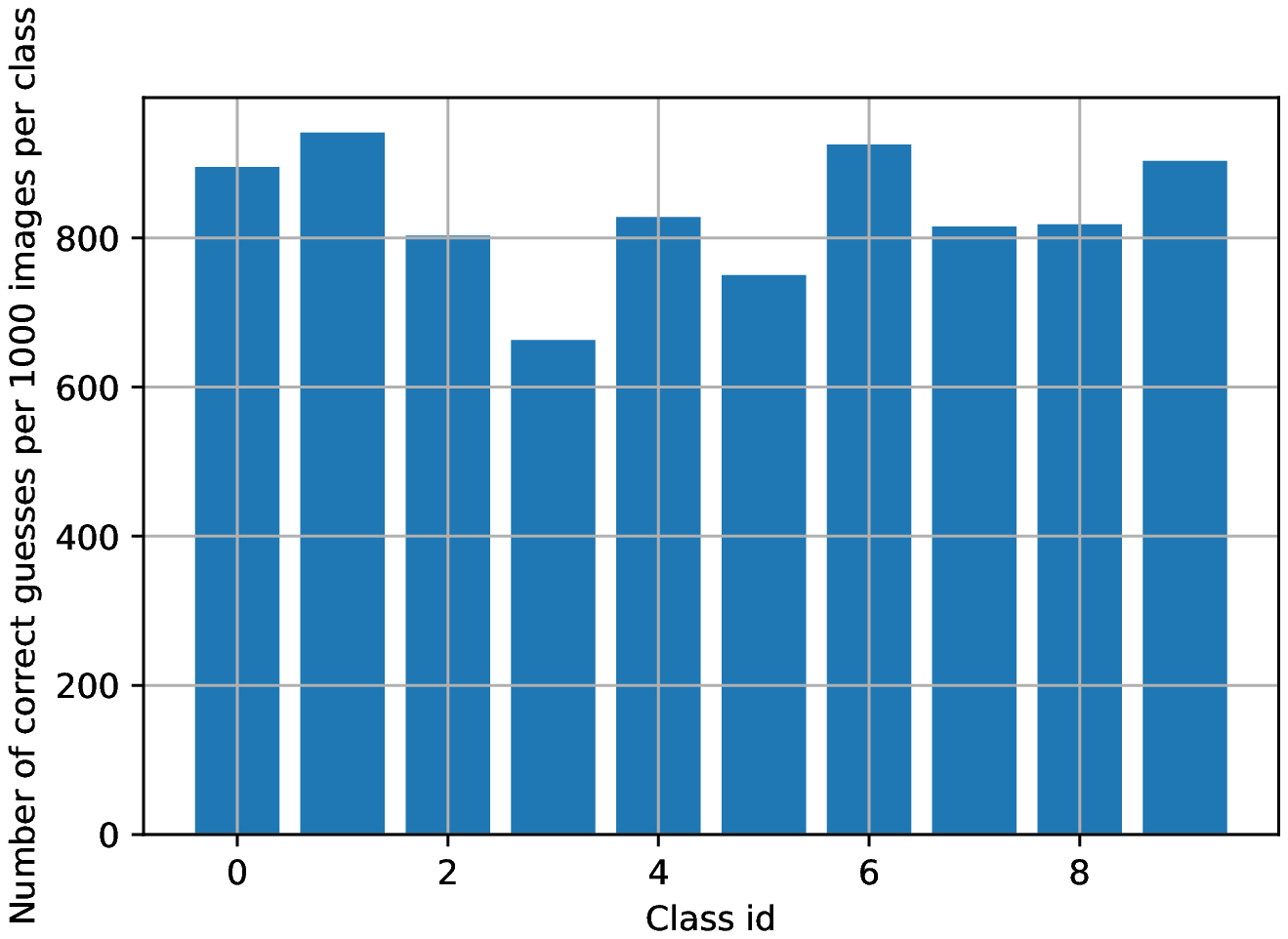}
	\end{minipage}}
 \hfill
  \subfloat[DAS]{
	\begin{minipage}[c][1\width]{
	   0.29\textwidth}
	   \centering
	   \includegraphics[width=1\textwidth]{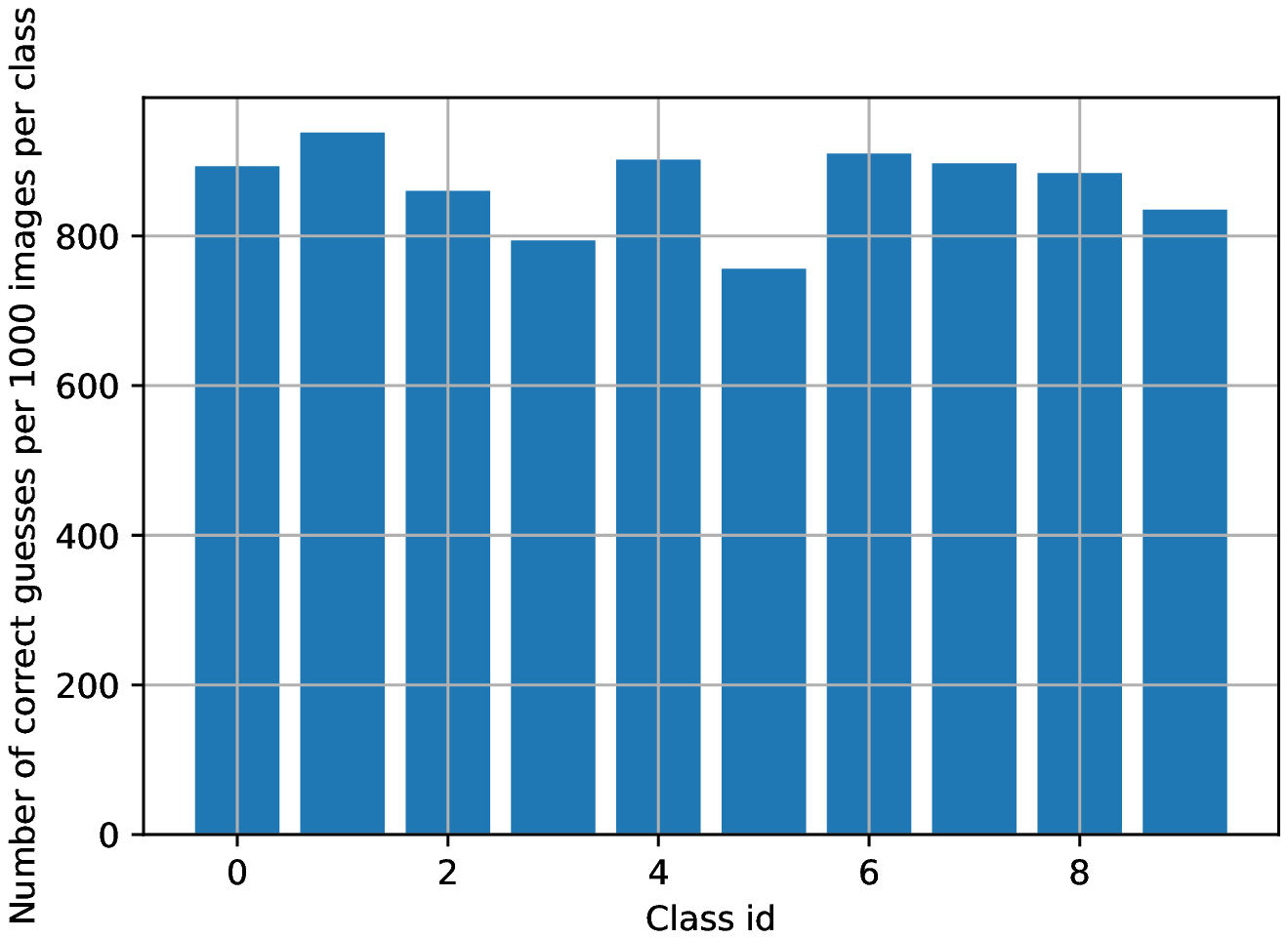}
	\end{minipage} }

 \hfill
 
\caption{Correct classified item/class for 10000 samples(1000 samples on each class)}
\vskip -0.2in
\end{figure}
\section{Discussion}
\subsection{Comparison between Dual Sampling and Coreset}
In Figure \ref{fig:1a}, DAS gave a preferable result compare with core-set selection in the beginning and surpassed the core-set method after it reached around 5000 instances.  While most other AL methods except core-set tend to perform well in the beginning and get closer to random selection in the end, DAS seems to behave oppositely. Since DAS is depended on the output of the models, it is likely that these models need to be stable and converge in order for DAS to work. In terms of computational time, DAS is both faster and highly parallelizable compare with core-set selection since the networks can get trained parallelly, i.e., if given enough computational power, DAS can theoretically be nearly as fast as random selection.
\subsection{Distribution of label and accuracy per class}
In Figure 2, the result of DAS shown a very unbalanced distribution of classes within the selected dataset, despite the fact that both random sampling and Core-set sampling followed the original balanced distribution of the Cifar-10. Besides that, DAS gave a substantial difference in number between class 1, 320 images of \textbf{ships}, and class 4, 1802 images of \textbf{cats}. By looking at Figure 3a and 3b, in terms of accuracy, cats seem to be the most challenging class in Cifar-10 while images of ships seem to be much easier to classify. Furthermore, on Figure 3c, while the model that used DAS selection had only be trained on around 300 of ship's images, it was able to achieve a comparable accuracy with the other results, which trained on nearly a thousand of pictures. It is also worth noting that in the case of DAS, the four classes in the middle got selected more often the other classes, and they are also the classes that have the lowest accuracy in both core-set and random selection. By looking more closely at the pictures in Figure 4 and Figure 5, while the environment/background and object form are quite similar(sky and water) in the case of ship's pictures, the variation of background and form for cat's pictures is much more significant. This difference can explain why images of cats are more difficult for a model to learn to classify than images of ships.
\begin{figure}[ht]
\vskip -0.1in
	   \centering
	   \includegraphics[width=1\textwidth]{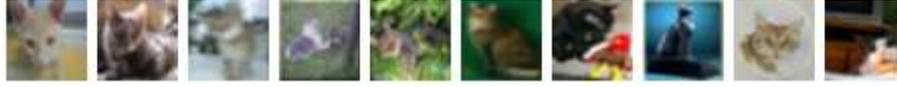}
  \caption{Pictures of cats from Cifar-10}
  \label{fig:3}
\end{figure}
\begin{figure}[ht]
\vskip -0.1in
	   \centering
	   \includegraphics[width=1\textwidth]{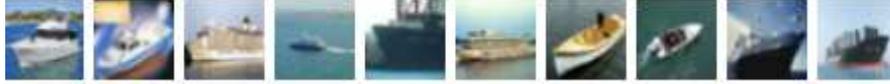}
  \caption{Pictures of ships from Cifar-10}
  \label{fig:4}
\vskip -0.4in
\end{figure}

\section{Conclusion}
In this paper, dual active sampling was introduced as a simple yet effective method in active learning. However, the method still needs to be tested on more data set to confirm its effectiveness. 
The method was able to give an impressive result on the Cifar-10 dataset with superior accuracy and computational time compared to core-set selection, which is found to be very interesting provided the simplicity of its implementation. By looking at the distribution of selected items per class and the respective accuracy, DAS has shown an impressive behavior. It was able to select more samples from the challenging classes while maintaining a comparable accuracy on the less challenging classes, with much less sampling effort. 

%
%
%
%

\end{document}